%% file: main.tex
\documentclass{applemlr}

\usepackage{amsmath}
\usepackage{enumerate}
\usepackage{algorithm}
\usepackage{algpseudocode}
\usepackage{amsfonts}
\usepackage{amsthm}
\usepackage{newtxtt}
\usepackage[noabbrev,nameinlink]{cleveref}
\usepackage{diagbox}
\usepackage{colortbl}
\usepackage{amssymb}
\usepackage{xspace}
\usepackage{wrapfig}
\usepackage{adjustbox}
\usepackage{tabularx}
\usepackage{booktabs}
\usepackage{mathtools}
\usepackage{tikz}
\usepackage{enumitem}
\usepackage{silence}
\usepackage{dsfont}
\usepackage[table]{xcolor}
\usepackage[dvipsnames]{xcolor}
\usepackage{multirow}
\usepackage{makecell}
\usepackage{xfakebold}
\usepackage{graphicx}
\usepackage{hepnames}
\usepackage{comment}
\usepackage{cuted}
\usepackage{url}
\usepackage{wrapfig}
\input{math_commands}

\definecolor{textgray}{HTML}{6E6E73}
\usetikzlibrary{positioning, calc}
\usetikzlibrary{decorations.pathmorphing}

\makeatletter
\patchcmd{\wrong@fontshape}{\@gobbletwo}{}{}{}
\makeatother
\numberwithin{equation}{section}
\tcbuselibrary{minted}
\setminted[python]{
  linenos,
  breaklines,
  fontsize=\footnotesize,
  xleftmargin=2em
}

\makeatletter
\AtBeginDocument{
  \urlstyle{sf}
  
}
\makeatother

\definecolor{light}{RGB}{125, 125, 125}
\crefname{tcb@cnt@pbox}{code}{code}
\Crefname{tcb@cnt@pbox}{Code}{Code}
\crefname{assumption}{assumption}{assumption}
\Crefname{assumption}{Assumption}{Assumptions}

\newtcolorbox[auto counter]{pbox}[2][]{
  colback=white,
  title=Code~\thetcbcounter: #2,
  #1,fonttitle=\sffamily,
  fontupper=\sffamily,
  arc=2pt,
  colframe=bgcolor,
  coltitle=fgcolor,
  colbacktitle=bgcolor,
  toptitle=0.25cm,
  bottomtitle=0.125cm
}

\makeatletter
\newcommand\applefootnote[1]{%
  \begingroup
  \renewcommand\thefootnote{}%
  \renewcommand\@makefntext[1]{\noindent##1}%
  \footnote{#1}%
  \addtocounter{footnote}{-1}%
  \endgroup
}
\makeatother

\definecolor{cverbbg}{gray}{0.90}

\newcommand*\model{{FAE}}


\title{One Layer Is Enough: Adapting Pretrained Visual Encoders for Image Generation}
\author{Yuan Gao}
\author{Chen Chen}
\author{Tianrong Chen}
\author{Jiatao Gu}

\affiliation{Apple}

\abstract{
\input{sec/0_abstract}
}

\metadata[Correspondence]{\sffamily Yuan Gao (\url{ygao65@apple.com}), Jiatao Gu (\url{jgu32@apple.com})}

\date{\sffamily\today}

\begin{document}

\maketitle

\input{sec/1_intro}

\input{sec/2_related_work}
\input{sec/3_method}
\input{sec/4_experiments}

\input{sec/5_ablations}

\input{sec/6_conclusion}

{
    \small
        }

\bibliographystyle{ieeenat_fullname}
\bibliography{main}

\appendix
\input{sec/7_appendix}

\applefootnote{ \textcolor{textgray}{\sffamily Apple and the Apple logo are trademarks of Apple Inc., registered in the U.S. and other countries and regions.}}

\end{document}

%% file: math_commands.tex
\usepackage{amsmath,amsfonts,bm}

\def\eqref#1{equation~\ref{#1}}

\def\1{\bm{1}}

\DeclareMathAlphabet{\mathsfit}{\encodingdefault}{\sfdefault}{m}{sl}
\SetMathAlphabet{\mathsfit}{bold}{\encodingdefault}{\sfdefault}{bx}{n}



%% file: sec/0_abstract.tex
Visual generative models (e.g., diffusion models) typically operate in compressed latent spaces to balance training efficiency and sample quality. In parallel, there has been growing interest in leveraging high-quality pre-trained visual representations—either by aligning them inside VAEs or directly within the generative model. However, adapting such representations remains challenging due to fundamental mismatches between understanding-oriented features and generation-friendly latent spaces. Representation encoders benefit from high-dimensional latents that capture diverse hypotheses for masked regions, whereas generative models favor low-dimensional latents that must faithfully preserve injected noise. This discrepancy has led prior work to rely on complex objectives and architectures.
In this work, we propose \textbf{\model{}} (Feature Auto-Encoder), a simple-yet-effective framework that adapts pre-trained visual representations into low-dimensional latents suitable for generation using as little as a single attention layer, while retaining sufficient information for both reconstruction and understanding. The key is to couple two separate deep decoders: one trained to reconstruct the original feature space, and a second that takes the reconstructed features as input for image generation.
\model{} is generic—it can be instantiated with a variety of self-supervised encoders (e.g., DINO, SigLIP) and plugged into two distinct generative families--diffusion models and normalizing flows. Across class-conditional and text-to-image benchmarks, \model{} achieves strong performance. For example, on ImageNet 256×256, our diffusion model with CFG attains a near–state-of-the-art FID of 1.29 (800 epochs) and 1.70 (80 epochs). Without CFG, \model{} reaches the state-of-the-art FID of \textbf{1.48} (800 epochs) and 2.08 (80 epochs), demonstrating both high quality and fast learning.

%% file: sec/1_intro.tex
\section{Introduction}
\label{sec:intro}

\begin{wrapfigure}{alignment}{0.5\textwidth}
\vspace{-20pt}
    \centering
    \includegraphics[width=\linewidth]{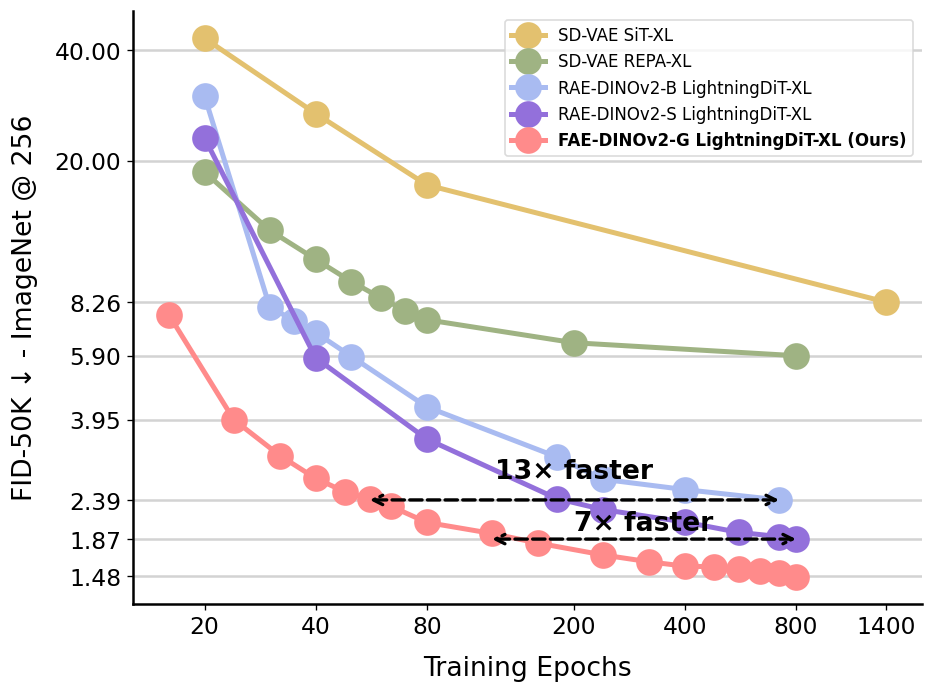}
    \caption{\textbf{Training convergence on ImageNet 256$\times$256.} FAE achieves strong sample quality and converges 7--13$\times$ faster than concurrent baselines~\citep{rae}.}
    \vspace{-10pt}
    \label{fig:comparison_rae}
\end{wrapfigure}

In past years, diffusion models \citep{ddpm, saharia2021image,ldm} have significantly advanced the quality and flexibility of visual generation, making them the dominant framework for producing high-resolution images and videos. A key recent change driving this progress is the integration of powerful pre-trained visual representations, typically obtained from large-scale self-supervised learning frameworks based on masked image prediction~\citep{dinov2,siglip2}. Such frameworks—exemplified by models like REPA~\citep{repa} and VA-VAE~\citep{vavae} utilize the rich semantic and structural information from the large models trained on unlabeled data.  When incorporated into the diffusion pipeline—either within the denoising process or in variational autoencoders (VAEs)—these representations substantially improve both training efficiency and generative fidelity. 

\begin{figure*}[t]
\centering
\includegraphics[width=\textwidth]{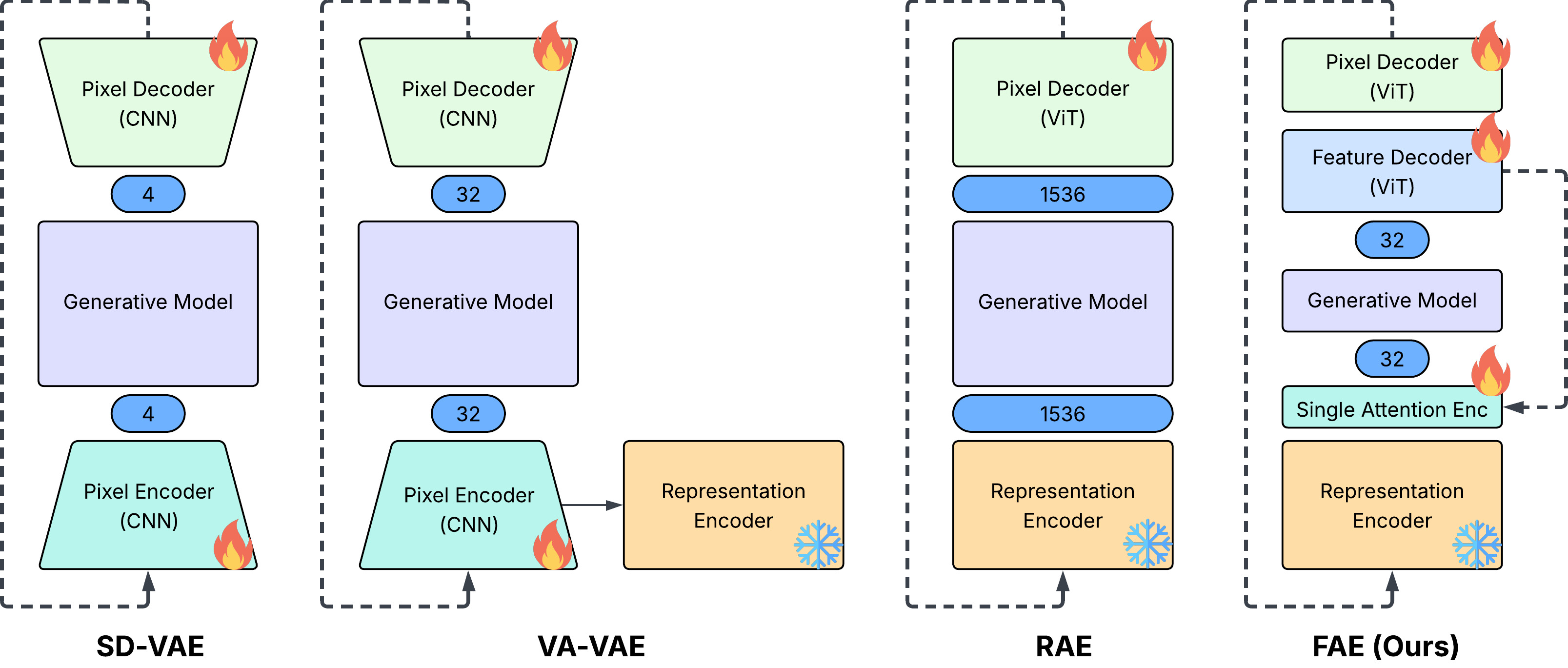}
\caption{
\label{fig:models} Comparison between standard VAE~\citep{ldm}, VA-VAE~\citep{vavae}, RAE~\citep{rae} and our proposed {\model}. The number shows the channel dimension of the generative modeling space. } 
\end{figure*}

Despite this progress, adapting pre-trained visual representations  to generative models  remains challenging due to the inherent mismatches between understanding-oriented representations and generation-friendly latent spaces.
Self-supervised models, in order to build up a hard task with the unlabelled data, masking and prediction tasks are used, not only in the image area but also text and audio. To capture the complicated distribution of  different possibilities of the masked regions, especially to simulate the distribution with simple embedding multiplication and softmax function, a large latent dimension is required.
In contrast, generation models  such as diffusion models and normalizing flow models, are often formulated as denoising processes, evolving a noisy input toward a clean signal through an iterative refinement process. In diffusion models, for instance, the input is perturbed by Gaussian noise and repeatedly denoised through multiple timesteps. To ensure trajectory stability throughout this process, the hidden representations must simultaneously encode the information of both the noised input and its clean predicted target. When the latent dimension is large, this dual encoding becomes more resource demanding, and the diffusion dynamics become sensitive to noise-level scheduling, often leading to instability or slower convergence.
Therefore, generative models favor compact, low-dimensional latent spaces, which make the denoising trajectory smoother, reduce the training burden, and preserve generative fidelity under limited model capacity.
This discrepancy often results in inefficiencies and necessitates complex architectural modifications when integrating pre-trained representations.

In this paper, we revisit this problem from a different perspective and ask: Is it truly necessary to preserve the high-dimensional structure of pre-trained visual representations when just zero-shot adapting them? In fact, although self-supervised models are trained on masked prediction tasks, the adaptation only involves unmasked inputs where the need for modeling diverse distribution is diminished. Instead, the goal is to leverage the rich semantics and spatial information from the pre-train features. 

Building on this insight, we introduce \textbf{\model{}} (\textbf{F}eature \textbf{A}uto-\textbf{E}ncoder), a simple yet effective framework that compresses the pre-train embedding  into a compact, generation-friendly space. We employ only a single attention layer followed by a linear projection to map the embeddings into a continuous low-dimensional code, and use a lightweight decoder to reconstruct the original features.
During experiments, we observed that the adaptation task is substantially weaker than the original self-supervised pre-training task; as a result, overly complex adapting frameworks tend to lose information from the pre-trained embeddings. Empirically maintaining a closer distance between the compressed code and the original embedding leads to higher reconstruction  quality.  Therefore, we adopt a minimal design, using only a single attention layer as the encoder to remove redundant global information shared across patch embeddings.

We validate our method by integrating it into existing diffusion frameworks, including SiT~\citep{sit} and LightningDiT~\citep{vavae} and normaling-flow based models (e.g., STARFlow~\citep{gu2025starflow}). On ImageNet 256×256 generation, with CFG, diffusion model using \model{} attains a near–state-of-the-art FID of 1.29 in 800 epochs and reaches an FID of 1.70 within only 80 training epochs. Without CFG, diffusion model using \model{} achieves a state-of-the-art FID of \textbf{1.48} in 800 epochs and reaches an FID of 2.08 with only 80 epochs, highlighting both its generation quality and its learning efficiency.

%% file: sec/2_related_work.tex
\section{Related Work}
\label{sec:related}

\noindent\textbf{Visual Representation Learning}~
Self-supervised learning (SSL) has become a cornerstone for learning general visual representations without manual labels. Early contrastive frameworks such as MoCo~\citep{moco} and SimCLR~\citep{simclr} maximized agreement between augmented views of the same image, while later non-contrastive approaches such as BYOL~\citep{byol} and DINO~\citep{dino} demonstrated that strong representations can emerge even without negative pairs. These methods trained large Vision Transformers (ViTs)~\citep{vit} to learn globally coherent, semantically rich feature spaces that transfer effectively to downstream tasks. 

More recent works have explored parameter-efficient ways to adapt such pretrained ViTs. Adapter-based and prompt-tuning methods—including AdaptFormer~\citep{adaptformer} and Visual Prompt Tuning (VPT)~\citep{vpt}—insert lightweight modules or learnable tokens into frozen backbones to tailor them to new domains with minimal fine-tuning overhead. However, most SSL-based adaptation studies focus on discriminative objectives—classification, segmentation, or retrieval—rather than generative modeling. Our work differs in that it repurposes a pretrained self-supervised ViT for visual generation, showing that a single attention layer suffices to bridge the gap between discriminative pretraining and generative diffusion training.

\noindent\textbf{Visual Generative Models}~
Diffusion models~\citep{ddpm, sde} have emerged as a dominant paradigm for image generation, achieving remarkable realism and diversity. LDM~\citep{ldm} further improved efficiency by performing denoising in a compressed latent space learned by a VAE.
Subsequent work has explored various architectural improvements to transformers, including DiT~\citep{dit} and SiT~\citep{sit}, which are designed for scalability and enhanced global context modeling. More recently, normalizing flow based models such as TARFlow~\citep{zhai2024normalizing} and STARFlows~\citep{gu2025starflow,gu2025starflowv} have emerged new type of generative models that become promising alternatives drastically different from standard diffusion models on visual generation.

\noindent\textbf{Representation Alignment}~
Aligning generative model representations with pretrained visual encoders has proven effective for stabilizing training and enhancing sample quality. REPA~\citep{repa} proposes to align noisy hidden features of diffusion transformers with clean image embeddings from a pretrained ViT, substantially accelerating convergence and improving fidelity. Follow-up work such as REPA-E\citep{repae} extends this idea by enforcing semantic consistency between latent tokens and image features throughout training. 
Recent analyses have emphasized the inherent tension between reconstruction quality and generation stability, underscoring the need to reconcile these competing objectives.
VA-VAE~\citep{vavae} highlights that high-dimensional latent spaces may favor reconstruction but hinder generative convergence, motivating strategies that align latent encoders with pretrained vision models.
Along this line, concurrent to our work, several studies have explored directly using pre-trained embeddings as tokenizer inputs to improve generation quality, including VFM-VAE~\citep{tiancibi} and RepTok~\citep{minggui}.
Contemporary work RAE~\citep{rae} directly adopts pre-trained embeddings as the diffusion latent space. While this avoids explicit alignment, it demands significant architectural changes to the generator (e.g., wider channels, additional heads) to accommodate high-dimensional feature maps. 

Our work complements these findings by reusing pretrained ViTs directly as the generative backbone rather than modifying the latent space, and demonstrates that lightweight adaptation via a single attention layer can yield stable and high-quality diffusion generation.
Where prior methods typically rely on external alignment losses or auxiliary projection heads to bridge discriminative and generative representations, our method internalizes the alignment process: by inserting a single-layer attention adapter into a pretrained ViT, we enable the model’s own attention mechanisms to reconcile discriminative and generative objectives. This design unifies feature reuse and representation alignment within a single, minimal architectural modification, reducing both training cost and overall system complexity.

%% file: sec/3_method.tex
\section{Motivation}
\label{sec.motivation}
A core challenge in adapting pre-trained visual representations for generative models is the inherent dimensional and functional mismatch between self-supervised understanding models and generation models.
Representation encoders~\citep{mae,dinov2,siglip2,clip}, whose performance typically improves with higher-dimensional feature spaces, naturally favor large embeddings. 
For example, Dino-V2-G~\citep{dinov2} has a dimension of $1,536$.
This has also become the de-facto choice in MLLMs~\citep{liu2023visual,bai2023qwen}. However, such high-dimensional representations are poorly suited for many generative models~\citep{ldm,dit,gu2025starflow} operating in a latent space. Unlike representation learning tasks, which only require encoding semantic information, generation tasks must accurately recover fine-scale details from noisy inputs, making the denoising process highly sensitive to the dimensionality and structure of the latent space, making it harder for the model to preserve and refine the injected noise, leading to instability and reduced sample quality. In normal cases, generation works in a much lower dimensional space, ranging from $4\sim 64$.

Existing methods tackle this mismatch from two main directions:
\textbf{(1) Feature alignment.}~
Methods such as REPA~\citep{repa} and VA-VAE~\citep{vavae} attempt to align the features of an external representation encoder with the generative model, either inside the generator or within a VAE. This typically requires carefully designed alignment losses and additional training stages. However, because the generator or VAE architecture is substantially different from the pre-trained encoder, such alignment inevitably discards information that is not immediately useful for generation, limiting the benefit of using the original representations.
\textbf{(2) Direct modeling.}~
More recently, RAE~\citep{rae} directly adopts pre-trained embeddings as the diffusion latent space. While this avoids explicit alignment, it demands significant architectural changes to the generator (e.g., wider channels, additional heads) to accommodate high-dimensional feature maps. As a result, the model design becomes tightly coupled to the embedding dimensionality, making it difficult to scale or transfer across different encoders.

This motivates us to seek a design that simultaneously
(1) keeps generation in a low-dimensional latent space, so that existing generative architectures can be reused without substantial ad-hoc modifications; and
(2) remains as close as possible to the representation-encoder feature space, so that we fully inherit the strengths of pre-trained features and can ideally recover their semantics from the learned latents, rather than relying on lossy alignment losses.

To this end, we introduce \textbf{\model{}}, a new Feature Auto-Encoding approach in which a lightweight encoder compresses high-dimensional representation features into a compact latent space tailored for generation. In the following, we formalize this design and describe the architecture and training objectives of \model{} in detail.

\section{Method}
\label{sec:method}

In this section, we introduce \model{}, a simple-yet-effective frame to bridge visual representation learning and generative models using feature-level autoencoders. 

\subsection{Single-Attention Encoder}
\label{subsec:single_attention_encoder}

We first train a feature encoder to compress frozen pre-trained embeddings into a low-dimensional latent space. To preserve as much information as possible from the large pre-trained model, we deliberately use a minimal encoder: \emph{a single self-attention layer followed by a linear projection} that maps pre-trained patch embeddings $\mathbf{x}$ to compact latents $\mathbf{z}$. This design reduces the parameter count and keeps the mapping close to the original feature space (see \Cref{fig:attention} in Appendix).

Importantly, the adaptation objective (feature reconstruction) is substantially weaker than the original self-supervised pre-training task (e.g., masked region prediction). Over-parameterized encoders tend to overfit this easier objective, effectively re-encoding features for reconstruction and discarding information that is not directly supervised. Empirically, we find that keeping the encoder shallow and the latents close to the pre-trained embeddings leads to higher reconstruction quality and better downstream understanding.
Notably, the self-attention layer is crucial: it operates across patch embeddings to remove redundant global information and redistribute capacity, whereas a purely linear projection acts independently on each feature dimension and cannot adaptively de-redundantize patch-wise information.
The original DINOv2 paper~\citep{dinov2} investigates several feature centering strategies, including moving-average centering after softmax and the Sinkhorn–Knopp algorithm. More recent work such as UniTok~\citep{unitok} also mentions the importance of attention mechanisms when compressing high-dimensional feature representations.
Despite different formulations, these methods and our single-attention encoder may share a common underlying principle of removing redundant global information .

Our ablation in \Cref{tab:vae_structure} confirms these observations: the single-attention encoder outperforms a purely linear encoder and yields significantly better reconstruction quality than a deep Transformer encoder.

\subsection{Double Decoder}
A central design in \model{} is to separate \emph{feature reconstruction} from \emph{image synthesis}. 
Given compressed latents $\mathbf{z}$, we first reconstruct the original representation space, and only then decode pixels from the reconstructed features. 
This ``double-decoder'' design lets us preserve the semantics of the frozen encoder while giving the pixel decoder the flexibility needed for high-fidelity image generation.

\noindent\textbf{Feature Decoder.}~
Starting from the compact latent $\mathbf{z}$, a 6-layer Transformer feature decoder reconstructs the original embedding $\hat{\mathbf{x}}$. 
Each layer uses the same hidden dimension as the corresponding pre-trained backbone (e.g., DINOv2) so that the reconstructed features live in a compatible representation space. 
We employ Rotary Positional Embedding (RoPE)~\citep{rope}, RMSNorm~\citep{rmsnorm}, and SwiGLU activations~\citep{swiglu}, which we find empirically improve stability and reconstruction quality.

The feature decoder is trained with a standard VAE objective consisting of an $\ell_2$ reconstruction term and a KL regularization term:
\begin{equation}
\mathcal{L}_\text{VAE} 
= \| \hat{\mathbf{x}} - \mathbf{x} \|_2^2 
+ \beta \, \mathrm{KL}\big(q(\mathbf{z} \mid \mathbf{x})\,\|\,p(\mathbf{z})\big).
\end{equation}
This encourages $\mathbf{z}$ to remain close to a simple prior while retaining enough information for accurate recovery of the pre-trained embeddings. Compare to prior work like VA-VAE \citep{vavae} and \cite{diffusion-regularization}, our loss is quite simple, it's just L2 Loss. This make our reconstructed results can be easily zero-shot adapted into downstream task trained on original pre-train embedding, which we will show the results in the following subsection.
In practice, we observe that high-quality feature reconstruction is a strong predictor of downstream generative and understanding performance.

\noindent\textbf{Pixel Decoder.}
On top of the reconstructed features $\hat{\mathbf{x}}$, we attach a ViT-L–based pixel decoder~\citep{vit} that maps them to RGB images. 
Conceptually, the feature decoder restores the ``language'' of the pre-trained encoder, and the pixel decoder learns to translate that language into pixels. 
By letting the pixel decoder operate on rich, semantically meaningful embeddings rather than raw latents, we simplify the generation problem and improve visual fidelity.
Following prior work~\citep{titok}, we train the pixel decoder with a combination of adversarial, perceptual, and reconstruction losses:
\begin{equation}
\mathcal{L}_\text{pix} 
= \lambda_\text{GAN}\mathcal{L}_\text{GAN} 
+ \lambda_\text{perc}\mathcal{L}_\text{perc} 
+ \lambda_\text{rec}\mathcal{L}_\text{rec}.
\end{equation}
Here, the adversarial term $\mathcal{L}_\text{GAN}$ encourages realistic textures and global coherence, the perceptual loss $\mathcal{L}_\text{perc}$ aligns high-level features with the ground-truth images, and the reconstruction term $\mathcal{L}_\text{rec}$ preserves low-level details. 
Together with the feature decoder, this yields a compact latent space that remains semantically faithful to the pre-trained encoder while supporting high-quality image synthesis.

We train the pixel decoder in two stages, entirely in the embedding space. 
In the first stage, we inject Gaussian noise into the frozen pre-trained embeddings and train the decoder to directly reconstruct images from these noisy features:
$
\tilde{\mathbf{x}} = \mathbf{x} + \epsilon, \quad \epsilon \sim \mathcal{N}\big(0, \sigma^2 I\big),
$
where we set $\sigma = 0.4$ for DINOv2 and scale it according to the norm of the pre-trained embeddings. 
This produces a \emph{Gaussian embedding decoder} that is robust to moderate perturbations in the representation space.

\noindent\textbf{Pixel Fine-Tuning.}
Once the first stage converges, we fine-tune the same pixel decoder on the reconstructed embeddings $\hat{\mathbf{x}}$ produced by the feature decoder. 
Because the decoder only operates on embedding space, the Gaussian embedding decoder can be reused across different variants of {\model} without architectural changes. 
Remarkably, even without fine-tuning on $\hat{\mathbf{x}}$, the Gaussian embedding decoder already achieves strong generation quality , indicating that our compressed latent space preserves the majority of the information in the original pre-trained embeddings.
We visualize the overall training stages in \Cref{fig:stages}.
The detailed parameters and pixel decoder reconstruction quality are available in the Appendix.

\subsection{Generative Model Training}
\label{subsec:gen_training}

Once the latent space is ready, we in parallel train generative models \emph{directly} on the \emph{low-dimensional, compact} latents $\mathbf{z}$. 
During this stage, only the frozen backbone encoder and the single-layer feature encoder are used to map images into latent space, making generator training both memory- and compute-efficient. 
This decoupling turns the latent space into a modular interface: as long as a model can predict or transform $\mathbf{z}$, it can be used as a generator without any change to its core architecture. 
In this work, we instantiate two representative models on top of the same \model{} latent space: SiT~\citep{sit}, a diffusion model, and STARFlow~\citep{gu2025starflow}, a normalizing flow. 
For both, we adopt the default parameterizations and training configurations from their original works, simply replacing their native latent representation with our learned $\mathbf{z}$, without additional architectural tricks or alignment losses.

\begin{figure}[t]
\centering
\includegraphics[width=0.8\linewidth]{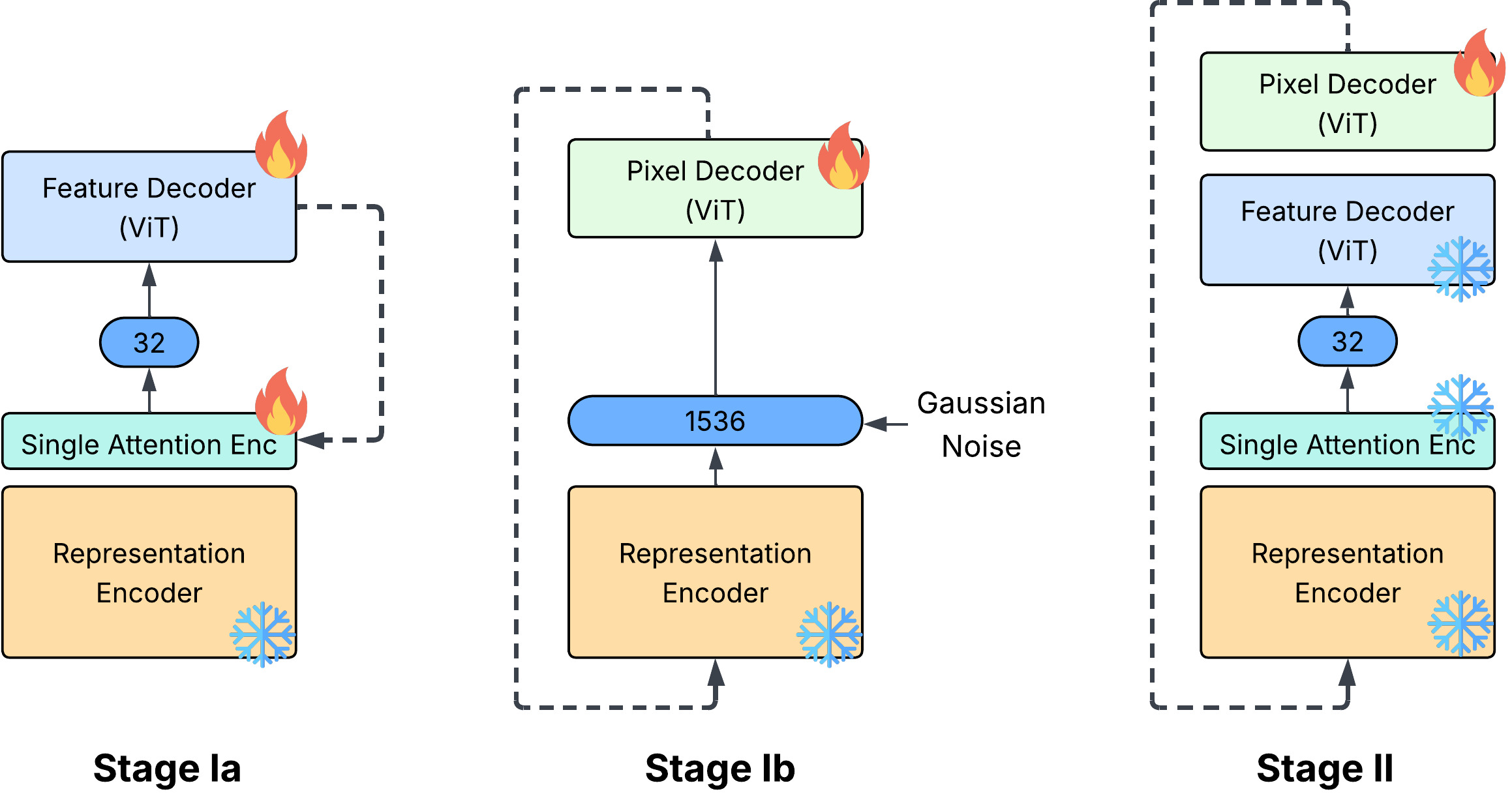}
\caption{\label{fig:stages} An illustration of Training Stages of \model{}. Stage Ia and Ib can be trained independently.} 
\end{figure}

\subsection{Semantic Preservation in the Latent Space}
\label{subsec:repr_preserving}

A key property of \model{} is that it can be directly applied to a variety of pre-trained visual representations (e.g., DINOv2~\citep{dinov2}, SigLIP~\citep{siglip2}) without architectural changes, while largely preserving their understanding capabilities, thanks to an explicit feature-decoder reconstruction objective that encourages the latent space to stay semantically close to the original embeddings rather than collapsing into a purely generative code.

We verify this by examining patch-wise similarity structure (see \Cref{fig:cat_sim} and \Cref{fig:panda_sim} in Appendix). After passing through \model{}, patches that are close in the original representation space remain close in our latent space, indicating that \model{} largely preserves the relational geometry of the pre-trained features.
Moreover, our latents retain the cross-image patch–matching behavior characteristic of DINOv2: semantically corresponding regions across different images (e.g., a player’s hand, an animal’s head) are still reliably matched using cosine similarity in the \model{}'s latent space (see \Cref{fig:cat_patch} and also \Cref{fig:bird_patch} \Cref{fig:elephant_patch} in Appendix). This suggests that \model{} preserves fine-grained, part-level semantics rather than only coarse global information.

\begin{figure}[htbp]
    \centering 
    \includegraphics[width=0.5\textwidth]{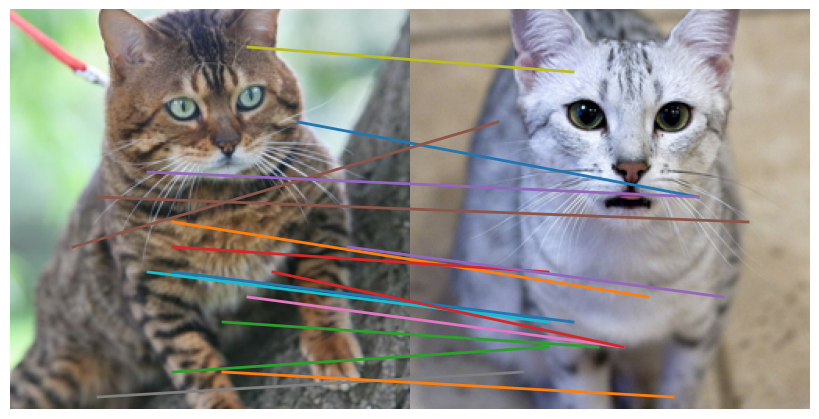}
    \caption{\label{fig:cat_patch}Matching across images.
    We match patch-level \model{} features between images from different images that share similar semantic information. This exhibits the ability of our model to understand relations between similar parts of different objects.
}
\end{figure}

\begin{figure}[htbp]
    \centering
    \input{sec/results/imagenet_probing}
    \captionof{table}{\label{tab:imagenet_probing} ImageNet Linear Probing top-1 accuracy comparison for FAE and different DinoV2 variants (all at 224 resolution).}
\end{figure}

%% file: sec/results/imagenet_probing.tex
\centering
\small
\begin{tabular}{lcc}
\toprule
\textbf{Model} & \textbf{Res} & \textbf{ImageNet top-1} \\
\midrule
DinoV2-S/14 distilled & 224 & 80.80\% \\
DinoV2-B/14 distilled & 224 & 84.40\% \\
DinoV2-L/14 distilled & 224 & 86.50\% \\
DinoV2-g/14 & 224 & 87.00\% \\
\midrule
{\model} (DinoV2-g/14) & 224 & 86.17\% \\
\bottomrule
\end{tabular}%

%% file: sec/4_experiments.tex
\section{Experiments}
\label{sec:expt}

We evaluate our proposed method on two standard generation benchmarks: class-conditional image generation on ImageNet-1K \citep{imagenet} and text-to-image generation trained on CC12M \citep{cc12m} and evaluated on MS-COCO \citep{mscoco}.
Our experiments show that the proposed approach substantially accelerates training convergence while improving overall generation quality.
To further demonstrate the generality of our framework, we also apply it to the STARFlow \citep{gu2025starflow} training paradigm and observe consistent improvements.
In addition, we investigate whether the learned latent representations preserve strong semantic understanding capabilities by performing zero-shot adaptation on common downstream tasks, including ImageNet-1K linear probing and MS-COCO image–text retrieval.

\input{sec/4_1_imagenet}
\input{sec/4_2_t2i}

\input{sec/4_3_tarflow}
\input{sec/4_4_understanding}

%% file: sec/4_1_imagenet.tex
\subsection{Class-conditional Image Generation}
\input{sec/results/main_results}

\textbf{Implementation details.} 
We processed images from ImageNet into resolution of 256×256 following ADM \citep{adm}. 
Then each image is then encoded into a compressed vector of shape 16×16×32 using {\model}.
For pre-train embedding, we explored DinoV2 \citep{dinov2}, we use a batch size of 1024 for training VAE. 
For latent diffusion model, we explored SiT following \cite{sit}'s setting and LightningDit following \cite{vavae}'s setting, we use the XL model size from these papers. To ensure a fair comparison with DiTs and SiTs, we consistently use a batch size of 512 during training.

\noindent\textbf{Evaluation and results.} 
For generation without CFG \citep{cfg}, we use SDE \citep{sde} and runs 250 steps. We achieve an FID score of 2.08 in 80 epochs and an FID score of 1.48 in 800 epochs. 
For generation with CFG, we use ODE and runs 250 steps. We achieve an  FID score of 1.70 in 80 epochs, and an FID score of 1.29 in 800 epochs.
Detailed CFG and timesteps shifts are available in the appendix.
Experiments on ImageNet demonstrate that our method can achieves state-of-the-art (SOTA) performance on image generation without CFG.
And the CFG results also got improved comparing to VA-VAE. 
We also provide some examples for the class conditioned generation in \Cref{fig:imagenet_final}.
Detailed hyper-parameter settings are provided in Appendix \Cref{sec:hyp_parameters}.

\begin{figure*}[htbp]
\centering
\includegraphics[width=\linewidth]{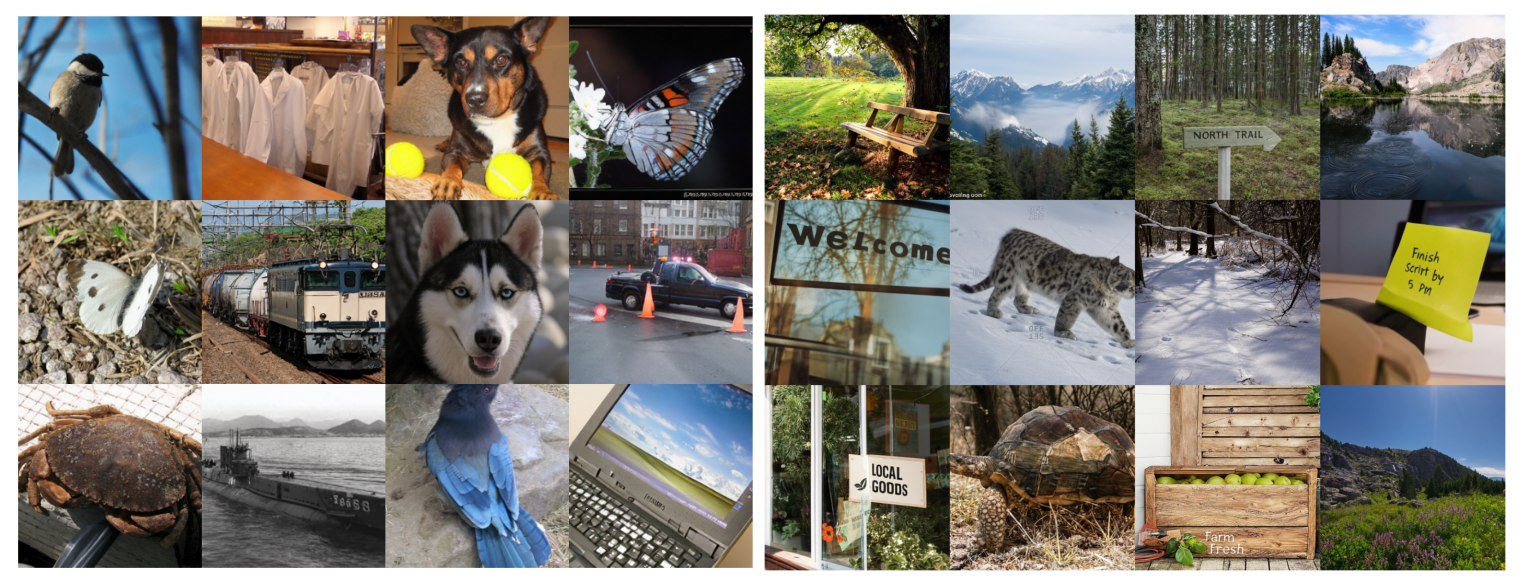}
\caption{\label{fig:imagenet_final} Random samples of ImageNet 256x256 and \label{fig:t2i_final} Text-to-Images using diffusion models.} 
\end{figure*}

%% file: sec/results/main_results.tex
\begin{table*}[t]
\centering
\resizebox{\textwidth}{!}{%
\begin{tabular}{l|c|c|ccccc|ccccc}
\toprule
\textbf{Method} &
\textbf{Training Epochs} &
\textbf{\#Params} &
\multicolumn{5}{c|}{\textbf{Generation w/o CFG}} &
\multicolumn{5}{c}{\textbf{Generation w/ CFG}} \\
\cmidrule(lr){4-8}\cmidrule(lr){9-13}
 & 
 & 
 & \textbf{gFID} & \textbf{sFID} & \textbf{IS} & \textbf{Pre.} & \textbf{Rec.} 
 & \textbf{gFID} & \textbf{sFID} & \textbf{IS} & \textbf{Pre.} & \textbf{Rec.} \\
\midrule

\multicolumn{13}{c}{\textit{\textbf{AutoRegressive (AR)}}} \\
\midrule
MaskGIT~\citep{maskgit}     & 555  & 227M & 6.18 & --   & 182.1 & 0.80 & 0.51 & --   & --   & --   & --   & --   \\
LlamaGen~\citep{llamagen}   & 300  & 3.1B & 9.38 & 8.24 & 112.9 & 0.69 & 0.67 & 2.18 & 5.97 & 263.3 & 0.81 & 0.58 \\
VAR~\citep{var}             & 350  & 2.0B & --   & --   & --    & --   & --   & 1.80 & --   & 365.4 & 0.83 & 0.57 \\
MagViT-v2~\citep{magvitv2}  & 1080 & 307M & 3.65 & --   & 200.5 & --   & --   & 1.78 & --   & 319.4 & --   & --   \\
MAR~\citep{mar}             & 800  & 945M & 2.35 & --   & 227.8 & 0.79 & 0.62 & 1.55 & --   & 303.7 & 0.81 & 0.62 \\
\midrule

\multicolumn{13}{c}{\textit{\textbf{Latent Diffusion Models}}} \\
\midrule
MaskDiT~\citep{maskdit}     & 1600 & 675M & 5.69 & 10.34 & 177.9 & 0.74 & 0.60 & 2.28 & 5.67 & 276.6 & 0.80 & 0.61 \\
DiT~\citep{dit}             & 1400 & 675M & 9.62 & 6.85  & 121.5 & 0.67 & 0.67 & 2.27 & 4.60 & 278.2 & \textbf{0.83} & 0.57 \\
SiT~\citep{sit}             & 1400 & 675M & 8.61 & 6.32  & 131.7 & 0.68 & 0.67 & 2.06 & 4.50 & 270.3 & 0.82 & 0.59 \\
FasterDiT~\citep{fasterdit} & 400  & 675M & 7.91 & 5.45  & 131.3 & 0.67 & \textbf{0.69} & 2.03 & 4.63 & 264.0 & 0.81 & 0.60 \\
MDT~\citep{mdt}             & 1300 & 675M & 6.23 & 5.23  & 143.0 & 0.71 & 0.65 & 1.79 & 4.57 & 283.0 & 0.81 & 0.61 \\
MDTv2~\citep{mdtv2}         & 1080 & 675M & --   & --    & --    & --   & --   & 1.58 & 4.52 & \textbf{314.7} & 0.79 & 0.65 \\
REPA~\citep{repa}           & 800  & 675M & 5.90 & --    & --    & --   & --   & 1.42 & 4.70 & 305.7 & 0.80 & 0.65 \\

VA-VAE~\citep{vavae} & 64  & 675M &
5.14 & 4.22 & 130.2 & 0.76 & 0.62 &
2.11 & 4.16 & 252.3 & 0.81 & 0.58 \\
 & 800 & 675M &
2.17 & 4.36 & 205.6 & 0.77 & 0.65 &
1.35 & \textbf{4.15} & 295.3 & 0.79 & 0.65 \\

RAE (DiT-XL)~\citep{rae} & 800 & 676M &
1.87 & -- & 209.7 & 0.80 & 0.63 &
1.41 & -- & 309.4 & 0.80 & 0.63 \\

RAE (DiTDH-XL)  & 80 & 839M &
2.16 & -- & 214.8 & 0.82 & 0.59 &
-- & -- & -- & -- & -- \\

 & 800 & 839M &
1.51 & -- & \textbf{242.9} & 0.79 & 0.63 &
\textbf{1.13} & -- & 262.6 & 0.78 & \textbf{0.67} \\

\midrule

\rowcolor{blue!10}
\textbf{{\model}} & 64 & 675M &
2.55 & 4.37 & 189.9 & 0.82 & 0.58 &
2.01 & 4.39 & 250.3 & 0.83 & 0.59 \\	
\rowcolor{blue!10}
 & 80 & 675M &
2.39 & 4.38 &	192.8 & 0.82	& 0.59 &
1.92	& 4.35	& 249.6	& 0.83	& 0.59 \\
\rowcolor{blue!10}
 & 800 & 675M &
1.58 & 4.38 & 223.7 & 0.80 & 0.63 &
1.41 & 4.34 & 274.1 & 0.81 & 0.63 \\
\midrule
\rowcolor{blue!10}
\textbf{\model\ w/ Timestep Shift} & 64 & 675M &
2.34 & 4.38 & 206.6 & \textbf{0.83} & 0.58 &
1.87 & 4.39 & 241.1 & 0.82 & 0.59 \\
\rowcolor{blue!10}
 & 80 & 675M &
2.08	& \textbf{4.20}	& 207.6 &	0.82	& 0.59 &
1.70	& 4.33	& 243.8	& 0.82	& 0.61 \\

\rowcolor{blue!10}
 & 800 & 675M &
\textbf{1.48} &	4.24 & 239.8 &	0.81	& 0.63 &
1.29 & 4.32 & 268.0 & 0.80 & 0.64 \\

\bottomrule
\end{tabular}%
}
\caption{\label{tab:main_results}\textbf{Class-conditional Image Generation Performance on ImageNet 256$\times$256.}}
\end{table*}

%% file: sec/4_2_t2i.tex
\subsection{Text-to-Image Generation}

\textbf{Implementation details.} 
For pre-train embedding, we reused the DinoV2 Encoder from Imagenet and train an extra Siglip2~\citep{siglip2} Encoder on ImageNet.
For training latent diffusion, we only use CC12M dataset with 256x256 resolution and follow the data processing from starflow \citep{gu2025starflow}'s setting. We consistently use a batch size of 256 during training. For text tokenzier, we use t5-xl.
We compare our model based on DinoV2 and Siglip2 embeddings. We also add SD-VAE as baseline.

\noindent\textbf{Evaluation and results.}
We evaluate \model{} on MS-COCO~\citep{mscoco} following the data preprocessing protocol of U-ViT~\citep{uvit}. For sampling \emph{without} using CFG~\citep{cfg}, we use an SDE sampler~\citep{sde} with 250 steps and obtain an FID of 7.47 after 400 training epochs. When using CFG, we switch to an ODE sampler with 250 steps and further improve the FID to 6.90 at 400 epochs. Notably, these near–state-of-the-art results are achieved using only CC12M for pre-training, i.e., with significantly fewer images than typical web-scale text-to-image models.

\input{sec/results/t2i_generation}

We also provide qualitative text-to-image examples in \Cref{fig:t2i_final}, and more examples in Appendix \Cref{fig:extra_examples}. These samples are generated at $384\times384$ resolution using a SigLIP2-\model{} backbone and an MMDiT decoder with 2B parameters. The model produces visually coherent images that follow short text prompts, indicating good text–image alignment.

Overall, experiments on COCO show that our method attains competitive, near-SOTA image generation quality while using substantially less training data. Detailed hyper-parameter settings are provided in the Appendix \Cref{sec:hyp_parameters}.

%% file: sec/results/t2i_generation.tex
\begin{table*}[t]
    \centering
    \resizebox{\textwidth}{!}{
    \begin{tabular}{lcccc}
    \toprule
        Model & FID & Type & Training datasets & \#Params \\
    \midrule

    \arrayrulecolor{black}
        \quad DALL-E~\citep{ramesh2021zero} & $\sim$ 28 & Autoregressive & DALL-E dataset (250M) & 12B \\
        \quad CogView~\citep{ding2021cogview} & 27.1 & Autoregressive & Internal dataset (30M) & 4B\\
        \quad LAFITE~\citep{zhou2021lafite} & 26.94 & GAN & CC3M (3M) & 75M + 151M (TE)  \\
        \quad GLIDE~\citep{nichol2021glide} & 12.24 & Diffusion & DALL-E dataset (250M)& 3.5B + 1.5B (SR) \\
        \quad Make-A-Scene~\citep{gafni2022make} & 11.84 & Autoregressive & Union datasets (without MS-COCO) (35M)& 4B\\
        \quad DALL-E 2~\citep{ramesh2022hierarchical} & 10.39 & Diffusion & DALL-E dataset (250M)& 4.5B + 700M (SR)\\
        \quad Imagen~\citep{saharia2022photorealistic} & 7.27 & Diffusion & Internal dataset (460M) + LAION (400M) & 2B + 4.6B (TE) + 600M (SR) \\
        \quad Parti~\citep{yu2022scaling} & 7.23 & Autoregressive & LAION (400M) + FIT (400M) + JFT (4B) & 20B + 630M (AE) \\
        \quad Re-Imagen~\citep{chen2022re} & \textbf{6.88} & Diffusion & KNN-ImageText (50M) & 2.5B + 750M (SR) \\
        \midrule
        \quad SDVAE+T5 (w/o CFG) & 21.25 & Diffusion & CC12M & 604M \\
        \midrule
        \rowcolor{blue!10}
        \quad \model{} (SigLIPV2)+T5 (w/o CFG) & 7.57 & Diffusion & CC12M & 604M+ 514M (\model{}) \\
        \rowcolor{blue!10}
        \quad \model{} (SigLIPV2)+T5 (w/ CFG) & 7.11 & Diffusion & CC12M & 604M+ 514M (\model{}) \\
        \midrule
        \rowcolor{blue!10}
        \quad \model{} (DinoV2)+T5 (w/o CFG) & 7.47 & Diffusion & CC12M & 604M+ 514M (\model{}) \\
        \rowcolor{blue!10}
        \quad \model{} (DinoV2)+T5 (w/ CFG) & 6.90 & Diffusion & CC12M & 604M+ 514M (\model{}) \\
    \arrayrulecolor{black}\bottomrule

    \end{tabular}}
    \caption{\label{tab:fid_ms}FID results of different models on MS-COCO validation ($256 \times 256$). All the models are trained on external dataset and zero-shot evaluated on MS-COCO using 30K example.}
\end{table*}

%% file: sec/4_3_tarflow.tex
\subsection{Latent Normalizing Flows}
\begin{figure}[h]
    \centering
    \begin{subfigure}[b]{0.49\textwidth}
        \centering
        \includegraphics[width=\textwidth]{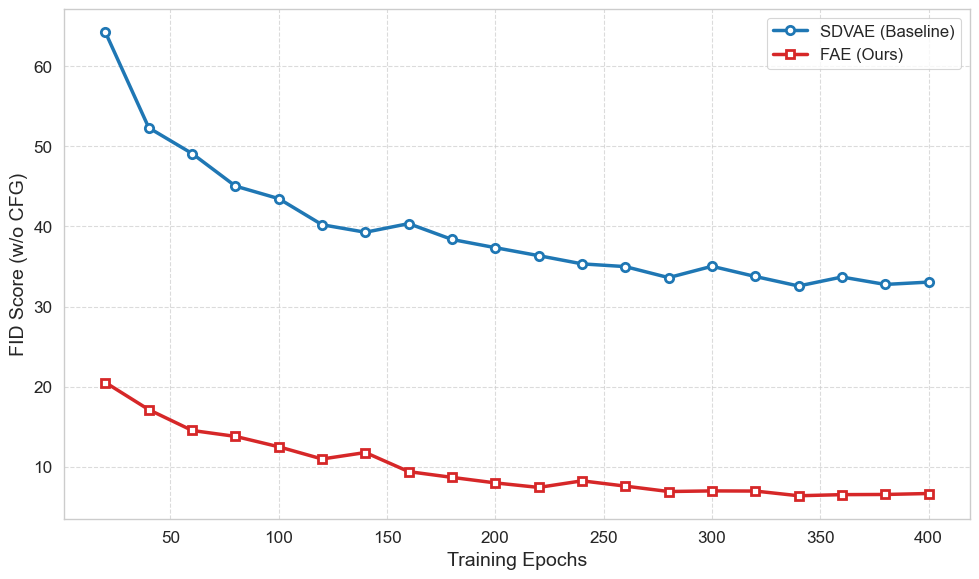}
        \caption{Results without CFG.}
        \label{fig:sub1}
    \end{subfigure}
    \hfill 
    \begin{subfigure}[b]{0.49\textwidth}
        \centering
        \includegraphics[width=\textwidth]{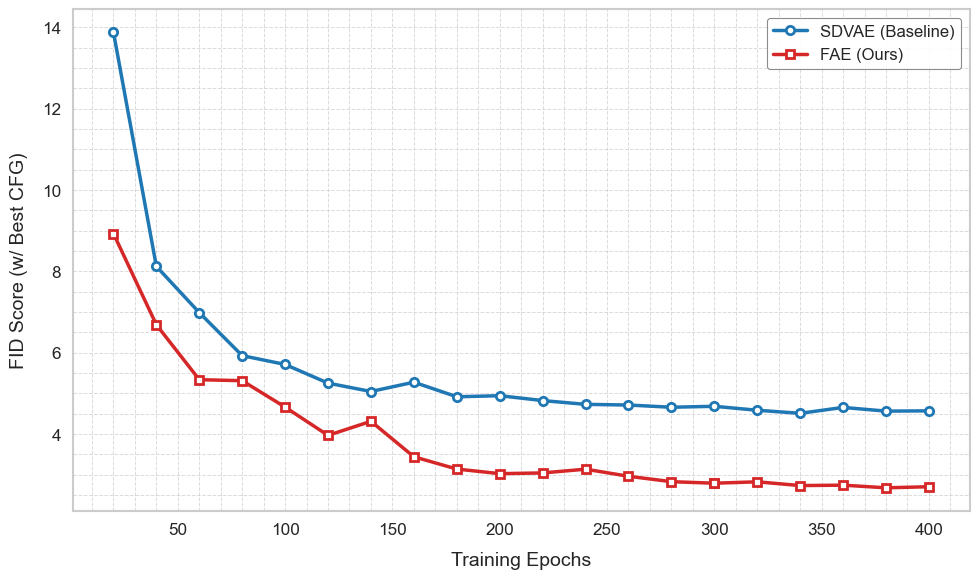}
        \caption{Results with CFG.}
        \label{fig:sub2}
    \end{subfigure}
    
    \caption{Comparison of STARFlow~\citep{gu2025starflow} with  SDVAE and the proposed FAE under the same settings.}
    \label{fig:both}
\end{figure}

We further validate the universality of \model{} by training a different family of latent generative model. Specifically, we instantiate STARFlow~\citep{gu2025starflow} -- an end-to-end latent normalizing flow generator that maps noise directly to \model{}’s latents, using the default configuration of 1.4B parameters and a standard SD-VAE baseline. The original STARFlow uses patch size 1 for SD-VAE latents, yielding sequences that are 4× longer than those of \model{}; for a fair comparison, we instead use patch size 2 for SD-VAE, resulting in a sequence length of 256 equivalent to \model{}.
As shown in \Cref{fig:both}, the SD-VAE baseline attains a FID of \textbf{4.51} under 400 training epoch, whereas the \model{}-based variant (DinoV2-g/14) achieves a FID of \textbf{2.67} and converges substantially faster for both guided and unguided scenarios. Additional visual results are provided in the Appendix.

%% file: sec/4_4_understanding.tex
\subsection{Image Understanding}
\label{sec:understanding}
We further validate the reconstructed embedding quality by zero-shot adapting it to the downstream tasks.
\noindent\textbf{Linear Probing.} 
We evaluate the DINOv2~\citep{dinov2} linear probing  using existing layer and weights from the origin model. We directly passed the {\model} reconstructed embedding to the layer. For data preprossing, we follow DINOv2. Espeically, we use the register version of {\model} (The generation results are similar to the one we listed in the main experiments, which is based on the non-register one.)
Results are shown in \Cref{tab:imagenet_probing}

\noindent\textbf{Text-Image Retrieval.} 
We evaluate the MS-COCO text-image dataset on COCO2014, following \href{https://github.com/google-research/big_vision/blob/main/big_vision/configs/proj/image_text/README_siglip2.md}{Siglip2}'s original setting.
We compute the text embedding using SigLip2~\citep{siglip2} text encoder. And compute the image embedding using image encoder and pass it to {\model} and use the reconstructed embedding. Then we compute the cosine similarity between the pair and find the Top-1.
Results are shown in \Cref{tab:coco_results}.

\input{sec/results/t2i_retrieval}

%% file: sec/results/t2i_retrieval.tex
\begin{table}[htbp]
\centering
\begin{tabular}{lcccc}
\hline
\textbf{Model} & \textbf{ViT} & \textbf{Res} & \textbf{COCO T$\rightarrow$I} & \textbf{COCO I$\rightarrow$T} \\
\hline
SigLIP2 & g-opt/16 & 256 & 55.45\% & 73.10\% \\
{\model} (SigLIP2) & g-opt/16 & 256 & 55.79\% & 72.94\% \\
\hline
\end{tabular}%
\caption{Text \& Image Retrieval on COCO dataset.}
\label{tab:coco_results}
\end{table}

%% file: sec/5_ablations.tex
\section{Ablation Study}
\label{sec:ablation}
In this section, we conduct several ablation experiments to examine the impact of each component in our single-layer adaptation framework.
For CFG-guidance, we grid search the CFG in 0.1 level and report the best results. All the CFG are enabled from t=0.7 to t=0.0 in the ablation experiments. For pixel decoder, we use a decoder fine-tuned with a 6 layer transformer reconstructed DinoV2 embedding.
Detailed experiment settings and results for each ablation study are available in \Cref{sec:hyp_parameters} in Appendix.

\noindent\textbf{FAE Model Structure.}
We evaluate the different FAE structure and architectural design in this subsection. We show that the shallower networks yield better generation quality and faster convergence.
Our single attention layer model outperforms both linear and 4/6 layer transformers in FID, while achieving comparable embedding-reconstruction similarity to deeper transformers—and substantially higher similarity than the linear baseline.
This suggests that a compact attention design not only accelerates optimization but also preserves semantic representation quality, potentially benefiting downstream zero-shot probing.
Results are available in \Cref{tab:vae_structure}
\input{sec/results/vae_structure}

\noindent\textbf{LDM Model Structure.}
We further analyze the model structure changes in the latent diffusion model. Starting from the base SiT architecture, we sequentially integrate SwiGLU, ROPE, and RMSNorm, resulting in a structure equivalent to LightningDiT.
Our conclusion is consistent with the Lightning paper, each component contributes positively to both convergence speed and overall generation quality, with the largest gains observed when all three are combined.
See \Cref{tab:sit_structure}
\input{sec/results/sit_structure}

\noindent\textbf{Token Dimension.}
We test VAE latent dimensions of 32, 48, and 64. After fine-tuning from the same decoder trained on Gaussian noise, the 64-dim model shows lower rFID than 48-dim and 32-dim variants
However, the final generation results indicate that the 32-dim setting achieves the best FID scores and IS scores, as well as the fastest convergence.
Notably, when time-shift is enabled (see below), the performance gap between different token dimensions narrows substantially.
Results are available in \Cref{tab:token_dim}
\input{sec/results/token_dim}

\noindent\textbf{Time Shift.}
Finally, we ablate the time-shift parameter in the diffusion process. By introducing time-shift, loss weighting and diffusion trajectory changes, thus effectively bridge the quality differences across VAE latent dimensions and significantly accelerate convergence. With time-shift, our model reaches state-of-the-art convergence speed and generation quality within only 64 epochs.
Results are available in appendix \Cref{tab:timeshift}

%% file: sec/results/vae_structure.tex
\begin{table}[t]
\centering
\small
\begin{tabular}{cccccc}
\hline
\textbf{Model} & \textbf{Linear Probing} & \textbf{CFG} & \textbf{64 Epochs} & \textbf{160 Epochs} & \textbf{320 Epochs} \\
\hline

\multirow{2}{*}{Single Attention}
  & 86.17\% & w/o & 2.98 & 2.27 & 1.98 \\
  &  & w/  & --     & 1.79 & 1.61 \\[2mm]

\multirow{2}{*}{Linear}
  & 85.74\% & w/o & 3.03 & 2.38 & 2.07 \\
  & & w/  & --     & 1.92 & 1.76 \\[2mm]

\multirow{2}{*}{6-Layer Transformer}
  & -- & w/o & 3.31 & 2.47 & 2.13 \\
  & -- & w/  & --     & 1.84 & 1.65 \\[2mm]

\multirow{2}{*}{\makecell[l]{Direct Predict\\ DinoV2}}
  & -- & w/o & 15.37 & 12.99 & 12.72 \\
  & -- & w/  & --      & 17.85 & 16.53 \\
\hline
\end{tabular}
\caption{Ablation results comparing different encoder structure.}
\label{tab:vae_structure}
\end{table}

%% file: sec/results/sit_structure.tex
\begin{table}[htbp]
\centering
\small
\begin{tabular}{p{3.5cm}cccc}
\hline
\textbf{Model} & \textbf{CFG} & \textbf{64 Epochs} & \textbf{160 Epochs} & \textbf{320 Epochs} \\
\hline

\multirow{2}{*}{SiT}
    & w/o & 2.98 & 2.27 & 1.98 \\
    & w/  & --     & 1.79 & 1.61 \\[2mm]

\multirow{2}{*}{SiT + SwiGLU}
    & w/o & 3.02 & 2.26 & 1.97 \\
    & w/  & --     & 1.75 & 1.60 \\[2mm]

\multirow{2}{*}{\makecell[l]{SiT + SwiGLU + ROPE}}
    & w/o & 2.86 & 2.182 & 1.89 \\
    & w/  & --     & 1.78 & 1.63 \\[2mm]

\multirow{2}{*}{\makecell[l]{SiT + SwiGLU +\\ ROPE + RMSNorm}}
    & w/o & 2.74 & 2.15 & 1.86 \\
    & w/  & --     & 1.71 & 1.55 \\
\hline
\end{tabular}
\caption{Ablation results comparing LDM model structure}
\label{tab:sit_structure}
\end{table}

%% file: sec/results/token_dim.tex
\begin{table}[htbp]
\centering
\small
\begin{tabular}{p{3.0cm}cccc}
\hline
\textbf{Model} & \textbf{CFG} & \textbf{64 Epochs} & \textbf{160 Epochs} & \textbf{320 Epochs} \\
\hline

\multirow{2}{*}{32-dim}
    & w/o & 2.67 & 2.02 & 1.76 \\
    & w/  & --     & 1.70 & 1.52 \\[2mm]

\multirow{2}{*}{48-dim}
    & w/o & 2.73 & 2.10 & 1.88 \\
    & w/  & --     & 1.73 & 1.56 \\[2mm]

\multirow{2}{*}{64-dim}
    & w/o & 2.86 & 2.25 & 1.99 \\
    & w/  & --     & 1.76 & 1.64 \\
\hline
\end{tabular}
\caption{Ablation results comparing different latent dimension.}
\label{tab:token_dim}
\end{table}

%% file: sec/6_conclusion.tex
\section{Conclusion}
We presented {\model}, a simple yet powerful framework for adapting high-quality self-supervised visual representations for generative modeling.
In contrast to prior approaches that rely on complex objectives or substantial architectural modifications to diffusion models, {\model} uses an extremely simple design: a single attention layer paired with two lightweight decoders. 
Across class-conditional and text-to-image benchmarks, {\model} demonstrates strong and efficient performance. On ImageNet 256×256, our diffusion model with CFG achieves a near–state-of-the-art FID of 1.29 with 800 training epochs and 1.70 with only 80 epochs. Without CFG, {\model} attains a state-of-the-art FID of 1.48 (800 epochs) and 2.08 (80 epochs), highlighting both its high sample quality and fast learning behavior.

Despite these promising results, {\model} still has limitations. Because the encoder is trained without an explicit image reconstruction loss, the rFID and tokenizer fidelity lag behind methods such as VA-VAE that directly optimize reconstruction quality.
Overall, {\model} provides a simple and general mechanism for leveraging pretrained vision encoders in generative modeling, offering a compelling balance between architectural minimalism, adaptability, and performance.

%% file: sec/7_appendix.tex
\clearpage
\setcounter{page}{1}

\section{FAE Encoder Structure}

\begin{figure}[htbp]
\centering
\includegraphics[width=0.5\linewidth]{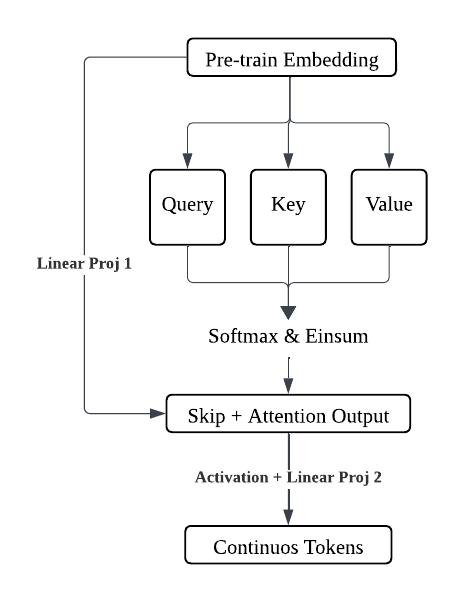}
\caption{
\label{fig:attention} Modified Attention} 
\end{figure}

We merge the consecutive linear layers in the attention module and use a larger per-head dimension for the encoder. The structure of our encoder is shown in \Cref{fig:attention}.

\newpage

\section{Ablation on FlowMatching Timesteps Shift}
Our ablation experiment demonstrates that applying Timesteps Shift accelerates convergence, mitigates the discrepancy across different latent token dimensions, and provides a modest improvement in the final FID score.
\input{sec/results/timeshift}

\section{rFID}
Because the encoder training is disentangled with the image reconstruction loss, our rFID and tokenizer reconstruction fidelity lag behind methods such as VA-VAE that directly optimize reconstruction quality.
\input{sec/results/rfid}

\onecolumn
\section{FAE Hyper Parameters}
\label{sec:hyp_parameters}
\input{sec/results/vae_hyp}
The MMDiT $384\times384$ and its {\model} encoder are only used for generating high quality examples provided in the paper.
The three-partitioned CFG are only used for generating Main Results. The main results uses timesteps shift=0.4. All ablation experiments use single cfg scale for $t=0.7\sim0.0$, and the cfg scale is grid searched in 0.1 fineness. The ablation experiments for FAE Model Structure are using SiT. The linear probing results are got from a separate encoder trained with DinoV2 register version with same parameters. And the ablation experiments for Token Dimension and Time Shift are using LightningDiT.

\onecolumn
\section{Patch Embedding Similarity Maps}
We compare the similarity between different patches inside single images.
For each triplet of visualizations, the first image shows the similarity map computed from the DINOv2 embeddings, while the second shows the corresponding similarity map derived from our {\model} latents. The third image displays the original image patch used as the query. The selected query patches are highlighted with a red rectangle, and darker colors in the similarity maps indicate higher similarity values.

\begin{figure*}[htbp]
    \centering 
    \includegraphics[width=0.8\textwidth]{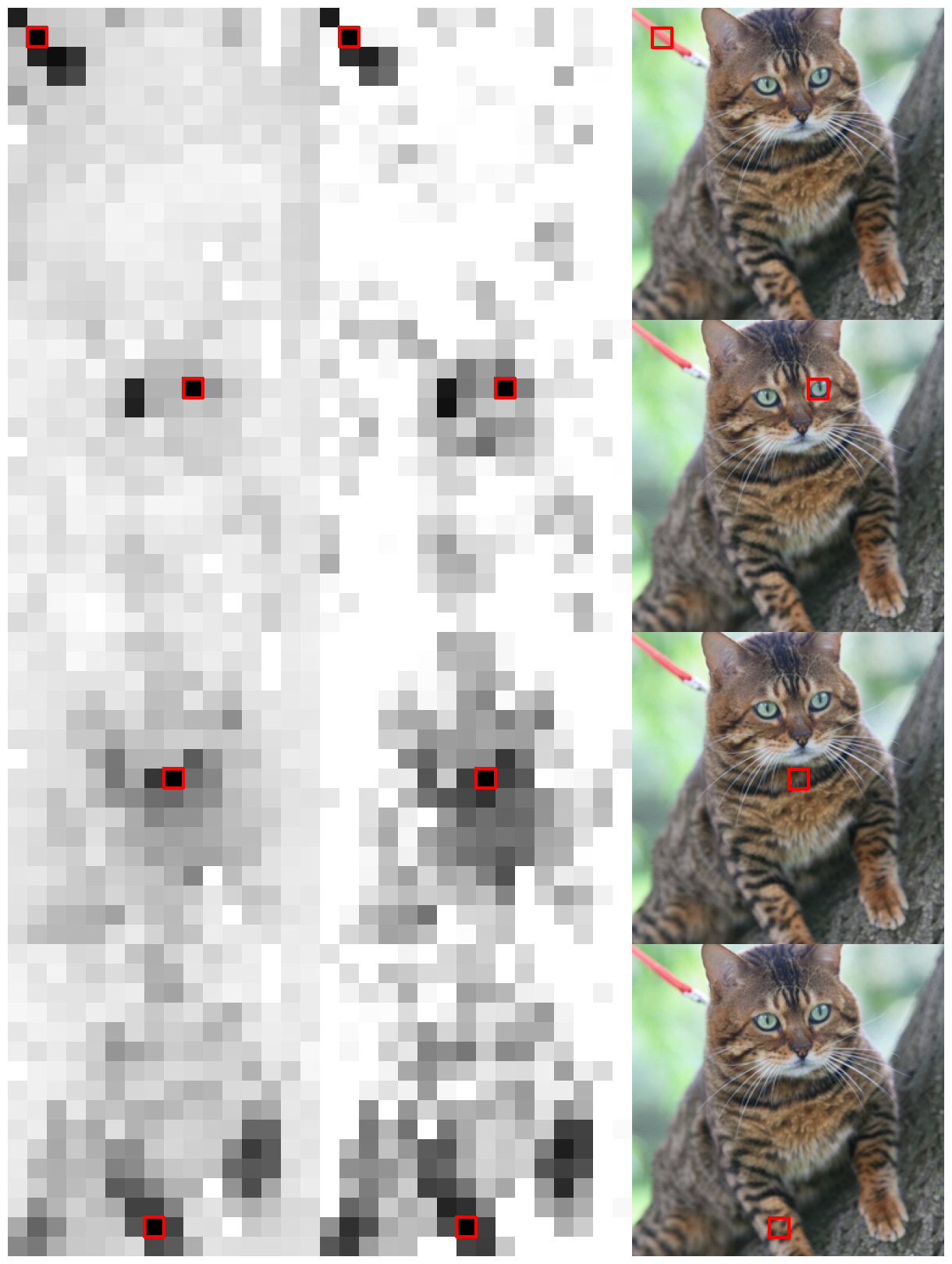}
    \caption{\label{fig:cat_sim}Similarity of a photo of cat. }
    \vspace{-10pt}
\end{figure*}

\begin{figure*}[htbp]
    \centering 
    \includegraphics[width=0.8\textwidth]{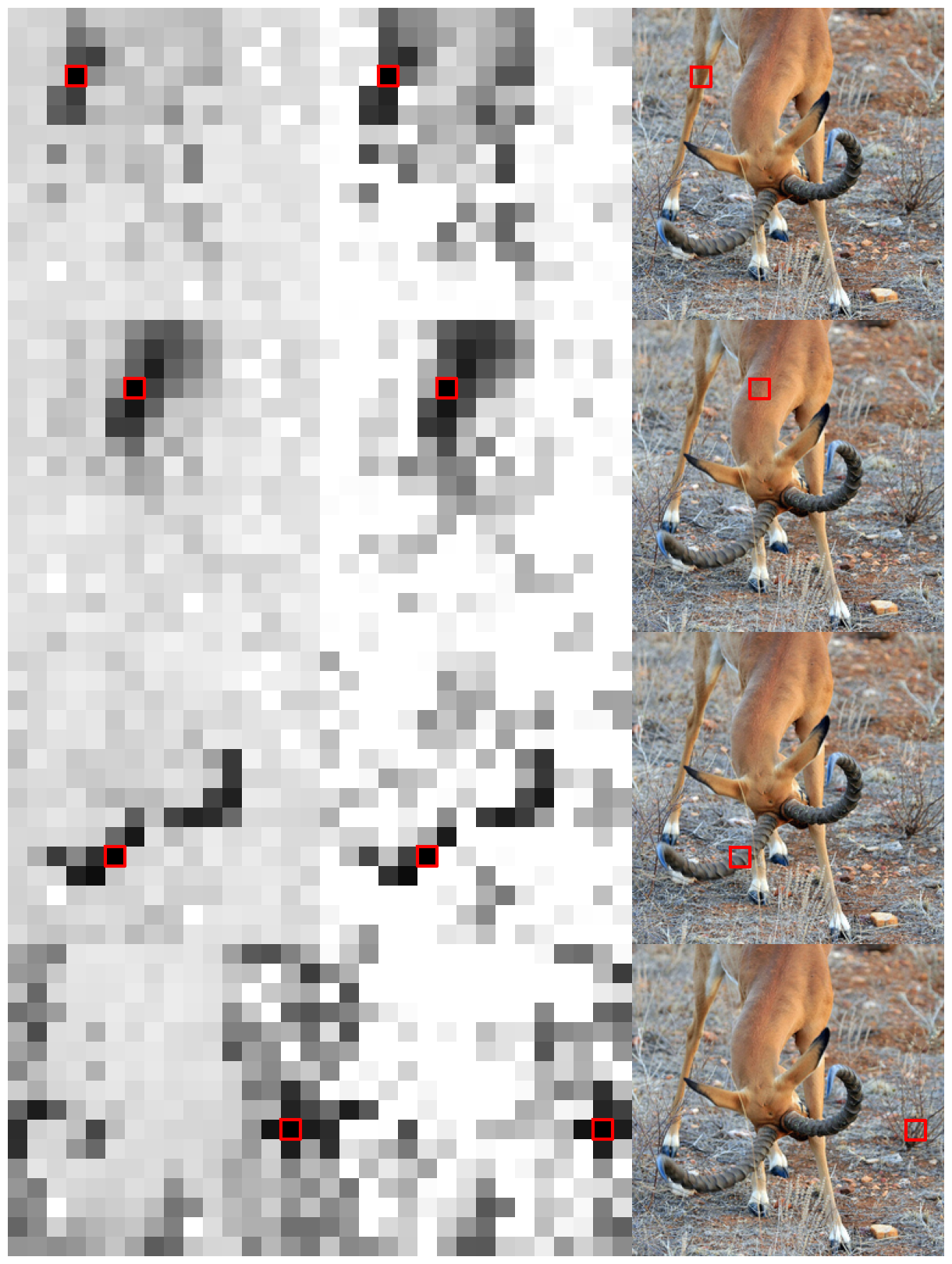}
    \caption{\label{fig:panda_sim} Similarity of a photo of impala.}
\end{figure*}

\onecolumn

\section{Matching most Similar patch pair across two images}
Our latents retain the cross-image patch–matching behavior characteristic of DINOv2: semantically corresponding regions across different images are still reliably matched using cosine similarity in the latent space.
This suggests that \model{} preserves fine-grained, part-level semantics rather than only coarse global information.

We first identify animal-related patches using K-Means clustering. From these, we randomly select patches in the first image and match each one to the patch in the second image with the highest cosine similarity. For each example, 16 patch pairs are selected and visualized.

\begin{figure}[htbp]
    \centering 
    \includegraphics[width=0.6\textwidth]{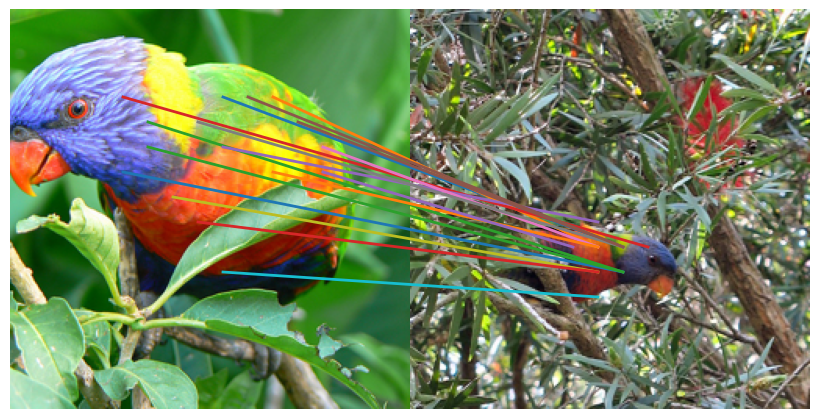}
    \caption{\label{fig:bird_patch} Matching most similar patch pair from two photo of bird.}
\end{figure}

\begin{figure}[htbp]
    \centering 
    \includegraphics[width=0.6\textwidth]{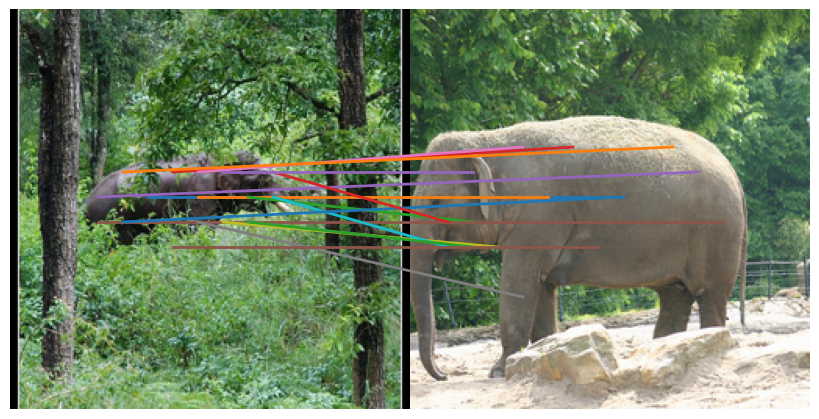}
    \caption{\label{fig:elephant_patch} Matching most similar patch pair from two photo of elephant.}
\end{figure}

\twocolumn

\onecolumn
\section{Text-to-Image Prompts}

The prompts for the text to image examples in \Cref{fig:t2i_final} are: \\
"a wooden bench under a large oak tree with warm sunlight streaming through branches and fallen leaves scattered below", \\
"a panoramic mountain ridge under soft morning clouds", \\
"a wooden arrow sign reading 'north trail' pointing into the woods",\\
"an alpine lake surrounded by steep cliffs, water perfectly still except for faint circular ripples, floating pollen creating tiny shimmering patterns on the surface",\\
"a window sticker reading 'welcome'",\\
"a snow leopard walking across snowy slope, faint pawprints trailing behind",\\
"a winter woodland with heavy snow draped asymmetrically across branches, faint animal tracks weaving between tree shadows under pale blue light",\\
"a sticky note attached to a monitor reading 'finish draft by 5 pm'",\\
"a small shop window sign reading 'local goods'",\\
"a tortoise lumbering through sunlit shrubs, shell etched with age patterns",\\
"a wooden produce crate stamped 'farm fresh' positioned in a sunlit garden shed beside tools",\\
"a rocky mountain meadow scattered with boulders and wildflowers under bright daylight"

\onecolumn
\section{STARFlow Examples}
\begin{figure*}[htbp]
\centering
\includegraphics[width=\textwidth]{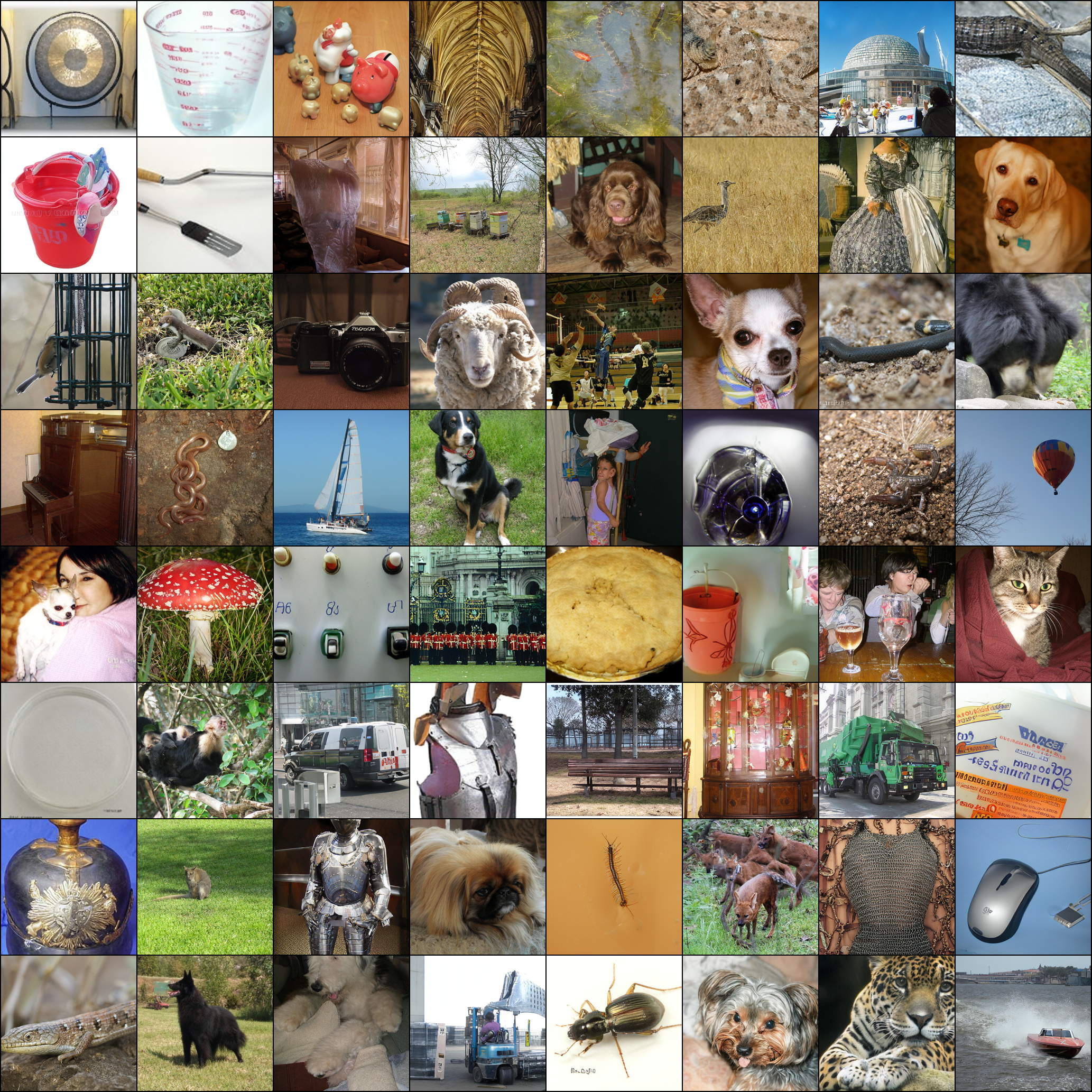}
\caption{\label{fig:starflow} Random samples of ImageNet 256x256 generated by STarFlow model.} 
\end{figure*}

\onecolumn
\section{Extra Examples from Siglip2 MMDiT $384\times384$ Model}
\begin{figure*}[b!]
\centering
\includegraphics[width=\textwidth]{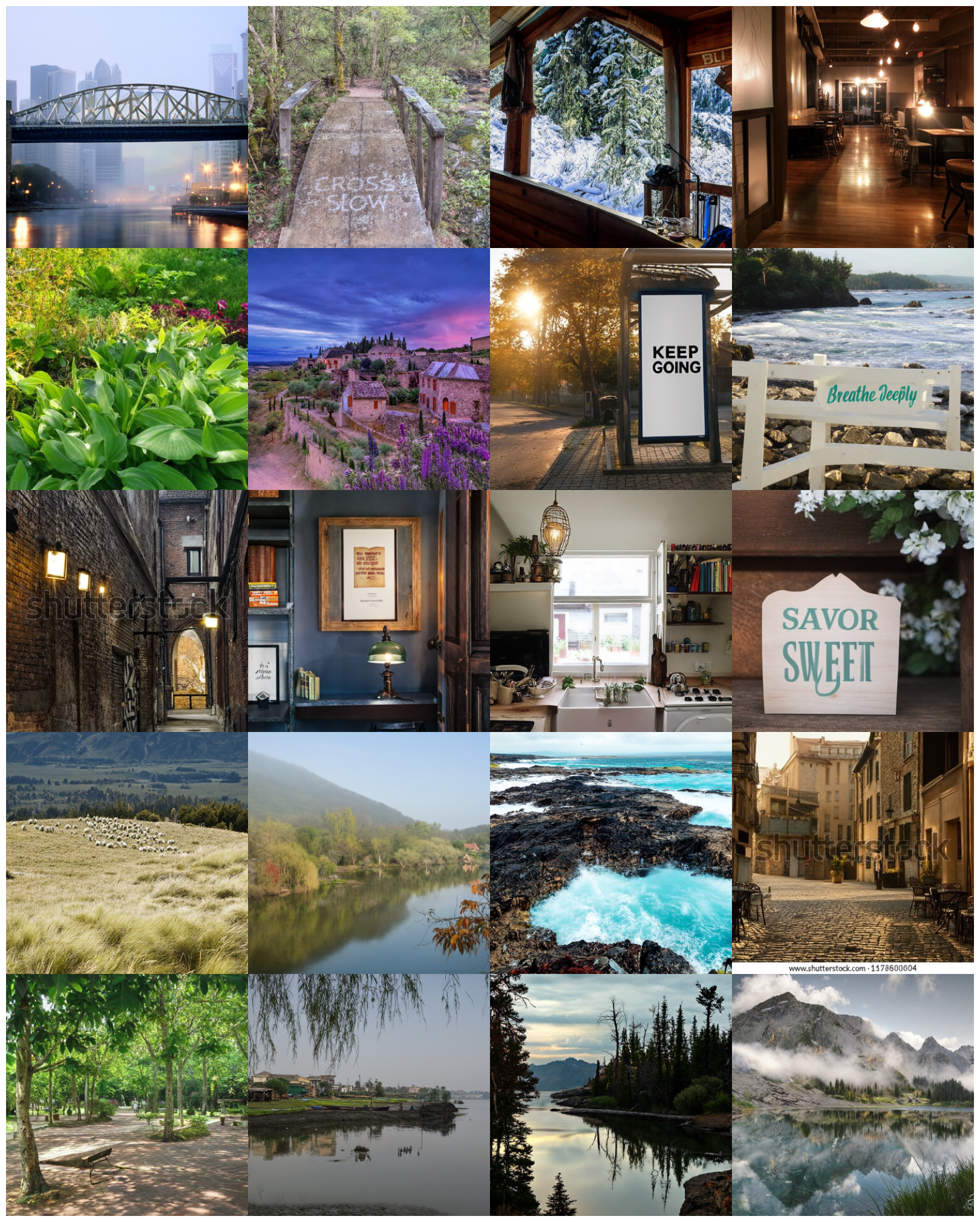}
\caption{\label{fig:extra_examples} Random samples of Siglip2 MMDiT $384\times384$ Model.}
\end{figure*}

\clearpage
The prompts for the text to image examples are: \\
"a foggy bridge spans a calm river reflecting muted city lights.",\\
"a forest bridge plank etched: ‘cross slow’. water murmurs beneath the boards.",\\
"a hiking lodge interior has a carved plaque: ‘rest; the mountains will wait.’ snow drifts past the windows.",\\
"a small café interior glows softly as candles flicker on wooden tables.",\\
"a quiet backyard garden with warm sun patterns filtering through leaves.",\\
"a hillside dotted with tiny cottages beneath a lavender evening sky.",\\
"a bus stop poster saying: ‘keep going’. golden morning light warms the street.",\\
"a seaside cabin porch sign saying: ‘breathe deeply’. waves pulse against the rocks below.",\\
"a lantern-lit alleyway glowing softly between old brick walls.",\\
"a quiet library alcove features a framed message: ‘seek answers, but also seek the calm between them.’ warm lamplight glows against tall wooden shelves.",\\
"a kitchen window frames sun-washed herbs, bowls, and warm shelves.",\\
"an orchard bench plaque reading: ‘savor sweet’. blossoms float in warm breezes.",\\
"a windy hillside covered in tall dry grass, each stalk catching light differently, distant sheep forming small irregular white clusters",\\
"a river winds beside a sleepy village, reflecting pale morning skies and drifting willow branches.",\\
"a stormy coastline where winds whip through rugged rocks and dark water.",\\
"a quiet european street lined with stone buildings glows under early dawn light as café chairs sit empty on cobblestones.",\\
"a quiet university quad filled with shaded benches and tree-lined paths.",\\
"a calm river flows beside a small town, reflecting the pale sky while fishermen prepare their nets and willow branches trail across the water’s surface.",\\
"a serene lake mirrors the surrounding pines and layered mountain ridges during early dawn.",\\
"a remote mountain lake reflects drifting clouds and jagged peaks.",\\

%% file: sec/results/timeshift.tex
\begin{table}[hbtp]
\centering
\begin{tabular}{p{3.8cm}cccc}
\hline
\textbf{Model} & \textbf{CFG} & \textbf{64 Epochs} & \textbf{160 Epochs} & \textbf{320 Epochs} \\
\hline

\multirow{2}{*}{\makecell[l]{32-dim, ts=0.7}}
    & w/o & 2.4087 & 1.9060 & 1.6836 \\
    & w/  & --     & 1.6786 & 1.5131 \\[2mm]

\multirow{2}{*}{\makecell[l]{32-dim, ts=0.5}}
    & w/o & 2.3233 & 1.8501 & 1.7086 \\
    & w/  & --     & 1.6735 & 1.5682 \\[2mm]

\multirow{2}{*}{\makecell[l]{32-dim, ts=0.3}}
    & w/o & 2.3220 & 1.8800 & 1.7743 \\
    & w/  & --     & 1.7125 & 1.6227 \\[2mm]

\multirow{2}{*}{\makecell[l]{48-dim, ts=0.5}}
    & w/o & 2.4329 & 1.9546 & 1.6952 \\
    & w/  & --     & 1.6797 & 1.5312 \\[2mm]

\multirow{2}{*}{\makecell[l]{48-dim, ts=0.3}}
    & w/o & 2.3599 & 1.9105 & 1.6911 \\
    & w/  & --     & 1.6694 & 1.5423 \\[2mm]

\multirow{2}{*}{\makecell[l]{64-dim, ts=0.2}}
    & w/o & 2.4398 & 1.9549 & 1.7581 \\
    & w/  & --     & 1.7563 & 1.5402 \\
\hline
\end{tabular}
\caption{Ablation results comparing different timesteps shift across different token dimension.}
\label{tab:timeshift}
\end{table}

%% file: sec/results/rfid.tex
\begin{table}[hbtp]
\centering
\small
\begin{tabular}{lcccc}
\hline
\textbf{} & \textbf{SD-VAE} & \textbf{VA-VAE} & \textbf{{\model} 32-dim} & \textbf{{\model} 64-dim} \\
\hline
\textbf{rFID} & 0.73 & 0.28 & 0.68 & 0.66 \\
\hline
\end{tabular}
\caption{\label{tab:rfid}Reconstruction rFID comparison.}
\end{table}

%% file: sec/results/vae_hyp.tex
\begin{table*}[h]
\centering
\resizebox{\textwidth}{!}{
\begin{tabular}{llccccccc}
\toprule
\textbf{Category} & \textbf{Field} &
\textbf{Encoder} & \textbf{Decoder} & \textbf{Pixel Decoder} &
\textbf{LDM} & \textbf{MMDiT} & \textbf{MMDiT 384x384} \\
\midrule

\multirow{8}{*}{Architecture}
 & Input dim.        
   & 16$\times$16$\times$1536 & 16$\times$16$\times$32 & 16$\times$16$\times$1536
   & 16$\times$16$\times$32 & 16$\times$16$\times$32 & 24$\times$24$\times$64 \\

 & Output dim.       
   & 16$\times$16$\times$64   & 16$\times$16$\times$1536 & 256$\times$256$\times$3
   & 16$\times$16$\times$32 & 16$\times$16$\times$32 & 24$\times$24$\times$64 \\

 & Hidden dim.        
   & 6144 & 1536 & 1024
   & 1152 & 1024 & 1536 \\

 & Num. layers       
   & 1 & 6 & 24
   & 28 & 16 & 24 \\

 & MLP Ratio         
   & -- & 4 & 4
   & 4 & 4 & 4 \\

 & Dim. per head      
   & 256 & 64 & 64
   & 72 & 64 & 64 \\

 & Num. heads        
   & 24 & 24 & 16
   & 16 & 16 & 24 \\

 & Total Params (M)  
   & 38.17 & 170.43 & 305.36
   & 675.26 & 603.46 & 2017.84 \\
\midrule

\multirow{7}{*}{Optimization}
 & Training iters   
   & \multicolumn{2}{c}{1M} & 1M
   & 2M & 1M & 1M \\

 & Batch size       
   & \multicolumn{2}{c}{1024} & 512
   & 512 & 512 & 512 \\

 & Optimizer        
   & \multicolumn{2}{c}{AdamW} & AdamW
   & AdamW & AdamW & AdamW \\

 & Peak LR          
   & \multicolumn{2}{c}{1e-4} & 1e-4
   & 1e-4 & 1e-4 & 1e-4 \\

 & LR Scheduler     
   & \multicolumn{2}{c}{Cosine} & Cosine
   & Constant & Constant & Constant \\

 & Warmup           
   & \multicolumn{2}{c}{1000} & 1000
   & -- & -- & -- \\

 & $(\beta_1,\beta_2)$
   & \multicolumn{2}{c}{(0.9, 0.999)} & (0.9,0.999)
   & (0.9,0.999) & (0.9,0.999) & (0.9,0.999) \\
\midrule

\multirow{8}{*}{Interpolants}
 & $\alpha_t$            
   & \multicolumn{2}{c}{--} & --
   & 1-t & 1-t & 1-t \\

 & $\sigma_t$            
   & \multicolumn{2}{c}{--} & --
   & t & t & t \\

 & $w_t$                 
   & \multicolumn{2}{c}{--} & --
   & $\sigma_t$ & $\sigma_t$ & $\sigma_t$ \\

 & Training objective    
   & \multicolumn{2}{c}{--} & --
   & v-prediction & v-prediction & v-prediction \\

 & Sampler               
   & \multicolumn{2}{c}{--} & --
   & \makecell{Euler-Maruyama (w/o CFG) \\ Euler (w/ CFG)}
   & \makecell{Euler-Maruyama (w/o CFG) \\ Euler (w/ CFG)}
   & \makecell{Euler-Maruyama (w/o CFG) \\ Euler (w/ CFG)} \\

 & Sampling steps        
   & \multicolumn{2}{c}{--} & --
   & 250 & 250 & 250 \\

 & Guidance              
   & \multicolumn{2}{c}{--} & --
   & \makecell{0.9 (t=1$\sim$0.9) \\ 2.5 (t=0.7$\sim$0)}
   & 1.5 (t=0.9$\sim$0)
   & 1.5 (t=0.9$\sim$0) \\
\bottomrule
\end{tabular}
}
\end{table*}